\documentclass{article}

\usepackage{PRIMEarxiv}

\usepackage[utf8]{inputenc} 
\usepackage[T1]{fontenc}    
\usepackage{hyperref}       
\usepackage{url}            
\usepackage{booktabs}       
\usepackage{amsfonts}       
\usepackage{nicefrac}      
\usepackage{microtype}      
\usepackage{lipsum}
\usepackage{fancyhdr}       
\usepackage{graphicx}       
\graphicspath{{media/}}     
\usepackage{placeins}

\usepackage{natbib}

\title{\textbf{Enhancing Math Learning in an LMS Using AI-Driven Question Recommendations}
}

\author{
  Justus Råmunddal \\
  \texttt{justus@hpkursen.se}
}

\begin{document}
\maketitle

\begin{abstract}
This paper presents an AI-driven approach to enhance math learning in a modern Learning Management System (LMS) by recommending similar math questions. Deep embeddings for math questions are generated using Meta's Llama-3.2-11B-Vision-Instruct model, and three recommendation methods---cosine similarity, Self-Organizing Maps (SOM), and Gaussian Mixture Models (GMM)---are applied to identify similar questions. User interaction data, including session durations, response times, and correctness, are used to evaluate the methods. Our findings suggest that while cosine similarity produces nearly identical question matches, SOM yields higher user satisfaction whereas GMM generally underperforms, indicating that introducing variety to a certain degree may enhance engagement and thereby potential learning outcomes until variety is no longer balanced reasonably, which our data about the implementations of all three methods demonstrate.

\end{abstract}

\section{Introduction}
Personalized learning is increasingly important in modern education, and recommendation systems are key to personalizing and optimizing the educational content presented to individual students. As highlighted by \citet{randieri_personalized_2024}, the integration of artificial intelligence into personalized learning is not only revolutionizing educational content delivery but is also redefining the traditional instructional model. The industry now views AI-powered adaptive systems as essential tools that tailor learning experiences to individual needs, thereby significantly enhancing student engagement and performance. Indeed, recent peer-reviewed studies further substantiate this trend. For example, \citet{halkiopoulos_leveraging_2024} demonstrate that AI-driven, cognitively informed adaptive assessment methods not only enrich personalized learning but also align it with modern pedagogical goals. Similarly, a systematic review on bioinspired adaptive learning algorithms underscores the rapid development and efficacy of these innovative approaches in creating dynamic, individualized educational environments \citep{yvonne_khomo_personalized_2025}. Together, these studies provide compelling evidence that advanced, AI-powered systems are pivotal in transforming personalized learning and adaptive instruction in today’s digital education landscape. Despite these advancements, a notable gap remains in the comparative analysis of distinct recommendation algorithms specifically designed for math content and in other content areas, particularly within learning management systems \citep{da_silva_systematic_2023}. 

In our LMS, we propose an AI-driven method for recommending math questions by leveraging deep embeddings generated using Meta's Llama-3.2-11B-Vision-Instruct model \citep{meta_llama-32-11b-vision-instructllama-32-11b-vision-instruct_2024}. We employ a straightforward cosine similarity approach that measures distance directly based on the vectors representing questions in the datasets, derived from the pretrained large language model (LLM). Furthermore, we explore SOM as a clustering method and later rank intracluster similarity by Euclidean distance to determine the most similar question. Last, we evaluate recommendations based on a GMM’s determined probability vector for a question compared to all other probability vectors of all other questions in the dataset, using KL divergence. Our evaluation tests the hypothesis that SOM and GMM can capture a richer semantic structure in the question space than cosine similarity alone.

This paper describes our methodology, experimental evaluation, and presents tables and visualizations that illustrate the performance and educational impact of these recommendation methods. The insights gained, aim to inform future adaptive system designs that can more effectively support personalized learning outcomes in math and beyond.

\section{Methodology}
\subsection{Deep Embedding Generation} \label{sec:deep-embedding-generation}
We generate embeddings for math questions in our datasets using Meta’s Llama-3.2-11B-Vision-Instruct model, available on Hugging Face \citep{meta_llama-32-11b-vision-instructllama-32-11b-vision-instruct_2024}. This model was chosen for its demonstrated capacity to yield semantically rich representations while balancing model size and performance. The survey by \citet{tao_llms_2024} indicates that large language models can serve as effective embedding models, with strategies such as direct prompting and data-centric tuning enhancing their embedding capabilities. Additionally, the paper discusses factors influencing the choice of embedding models, including performance versus efficiency and scaling laws. Our selection of an 11B-parameter model aligns with these insights, offering a robust trade-off: a 4096-dimensional embedding space capturing fine-grained semantics, without the prohibitive infrastructure needs of ultra-large models (e.g., GPT-4). Equally important, the selected model is multimodal, ingesting both text and images. Many math problems include diagrams, equations, or visual elements; integrating a vision--language encoder ensures these non-textual cues contribute to the embedding. This visual-text fusion aligns with recent findings that bridging math and vision in one model markedly improves problem understanding \citep{peng_multimath_2024}. In summary, Llama-3.2-11B-Vision-Instruct provides a high-fidelity embedding of each math question’s content, capturing textual context and any associated imagery, with an excellent embedding richness versus efficiency profile grounded in recent research on large multimodal models.

In our process, each math question is first preprocessed by undergoing basic text cleaning that strips each question string to ensure consistent formatting and to remove extra whitespace. Then, the Llama tokenizer from Hugging Face is applied to the math question’s text, and in some cases an accompanying image, which results in input data compatible with Meta’s Llama-3.2-11B-Vision-Instruct model. To obtain a semantically rich, single vector representation of each question, we first pass the tokenized question through the model, which transforms it into a series of vector representations, one for each token, in the final hidden layer. Then we extract the final hidden states for each question’s input tokens. After obtaining token-level hidden states from the model, we aggregate each question’s embedding via mean pooling (computing the arithmetic mean of all token embeddings). This method yields a single 4096-dimensional vector that encapsulates the full semantic and syntactic content of the question. Mean pooling is widely used in large language model applications for semantic similarity and sentence representation tasks due to its simplicity, robustness, and empirical effectiveness. Unlike max pooling, which may overemphasize a single salient token and amplify noise, mean pooling provides a balanced summary that reflects contributions from the entire input sequence \citep{tao_llms_2024}. Similarly, relying solely on a single-token embedding, such as a designated \texttt{[CLS]} token, can lead to suboptimal representations, as these are not always trained to encode sentence-level semantics without additional fine-tuning \citep{reimers_sentence-bert_2019}. In contrast, mean pooling consistently captures distributed contextual features across the input, resulting in stable and generalizable representations for downstream similarity comparisons. Given these properties of mean pooling, it offers an efficient and effective approach for generating question embeddings in our system without introducing model-specific tuning or complexity.

This approach allows us to compare math questions effectively by measuring distances in the high-dimensional embedding space. The embeddings serve as the foundation for our recommendation algorithms, enabling methods such as cosine similarity, Self-Organizing Maps (SOM), and Gaussian Mixture Models (GMM) to retrieve conceptually similar questions for users.

\subsection{Recommendation Algorithms} \label{sec:recommendation-algorithms}

We implement three distinct methods for recommending similar math questions: cosine similarity, Self-Organizing Maps (SOM) and Gaussian Mixture Models (GMM). Each method relies on the question embeddings generated in Section~\ref{sec:deep-embedding-generation} as a foundational data representation for distance calculations, model training, or probabilistic assignments. Hyperparameters were tuned based on preliminary experiments and domain considerations, all of which are specified in this section.

\subsubsection{Model-Specific Details}

Because the math questions across different SweSAT parts (XYZ, KVA, NOG, DTK) vary substantially in format and content, we train separate SOM and GMM models for each part and similarly constrain cosine similarity recommendations to remain within the same subject. At first glance, one might expect the embeddings alone to capture these differences, making separate boundaries unnecessary. However, in practice, some methods (e.g., GMM) operate on derived statistics like posterior probabilities of cluster membership rather than directly on the raw embeddings. These derived measures could potentially be sensitive to large cross-domain variation. Consequently, until further experimentation confirms each method's ability to handle multiple domains seamlessly, subject-specific boundaries remain the most reliable way to ensure contextually appropriate recommendations. In practice, the boundaries are implemented in our Django-system by mapping which dataset a question belongs to and then assigning the corresponding SOM and GMM for computation of the most similar datapoint from that same dataset.

\subsubsection{Cosine Similarity}

This method computes the cosine similarity between the embedding vectors from Section~\ref{sec:deep-embedding-generation} to find the nearest neighbors. Its simplicity and interpretability make it a useful baseline. However, its strict focus on exact similarity can sometimes lead to overly repetitive recommendations, especially in cases where a user's mistake may trigger a cascade of nearly identical questions.

\subsubsection{Self-Organizing Maps (SOM)}

SOM projects high-dimensional embeddings onto a two-dimensional grid while preserving the topological relationships of the original data. In our implementation, we configured the SOM with the following hyperparameters:

\begin{description}
\item[Grid Size] We chose a $5 \times 8$ grid. This size was selected because it offers a balance between resolution and generalization. A smaller grid might fail to capture the nuanced differences between math questions, whereas a larger grid might overfit to noise. Our manual evaluation shows that larger grids result in unreasonable question distributions, where the number of questions per output neuron was distributed unevenly. Choosing a moderate map size (e.g., 40 neurons in a $5 \times 8$ grid) is supported by recent findings on SOM performance. An appropriately sized SOM yields a good balance between capturing detail and avoiding over-sparsity \citep{guerin_survey_2025}. Too few neurons (too small a grid) will force dissimilar high-dimensional items into the same cluster, blurring important differences. Conversely, too many neurons (an overlarge grid) can produce excessively fine clusters that contain very few or zero samples, which adds noise and yields ``inactive'' neurons with no assigned data. In other words, small grids underfit (losing nuanced distinctions) while overly large grids overfit, fragmenting the data into sparsely populated micro-clusters \citep{kossakowski_analysis_2017}.

\item[Number of Epochs] The SOM was trained for 1000 epochs. This number of epochs allowed the network to adequately stabilize the mapping from the high-dimensional space to the 2D grid, ensuring that the topology is well preserved. Even though we chose to train the network for 1000 epochs, we observed acceptable maps already at 100 epochs. Researchers have found that on complex, high-dimensional data, SOMs often require on the order of hundreds to thousands of iterations to fully stabilize \citep{kossakowski_analysis_2017}. A 2024 clustering study reports that around 1000 epochs was needed for their SOM to yield optimal cluster separation, evidenced by the Davies--Bouldin index flattening out at the 1000-epoch mark. They note that the SOM achieved its best clustering performance (91\% accuracy, DBI $\sim 0.747$) at 1000 training iterations \citep{khotimah_enhancing_2024}.

\item[Learning Rate] Training started with a learning rate of 0.5, decaying linearly over the training period. This high initial rate quickly adjusts to major features in the data, and the decay allows for fine-tuning as the mapping converges. An initial learning rate of around 0.5 for SOM training is a common choice and is supported by contemporary experiments. \citet{wang_learning_2023}, configured their SOM with a learning rate of 0.5 in the initial phase, noting that this relatively high starting value helps the map quickly adjust to major features in the data (i.e., it promotes rapid organization in early epochs). Critically, the learning rate is then linearly decayed over time toward zero (or a small minimum) to allow fine-tuning.

\item[Intracluster Euclidean Distance] When predicting the most similar question in production, the algorithm treats each neuron in the SOM as a cluster and measures the intracluster Euclidean distance to identify the most similar question. In contrast, the straightforward cosine recommendation system treats all questions as part of the same cluster. So, to recommend a question, we find the Best Matching Unit (BMU) on the SOM for the query's embedding, then select the closest question within that neuron's cluster by Euclidean distance.
\end{description}

\begin{figure}[htbp]
  \centering
  \includegraphics[width=1\linewidth]{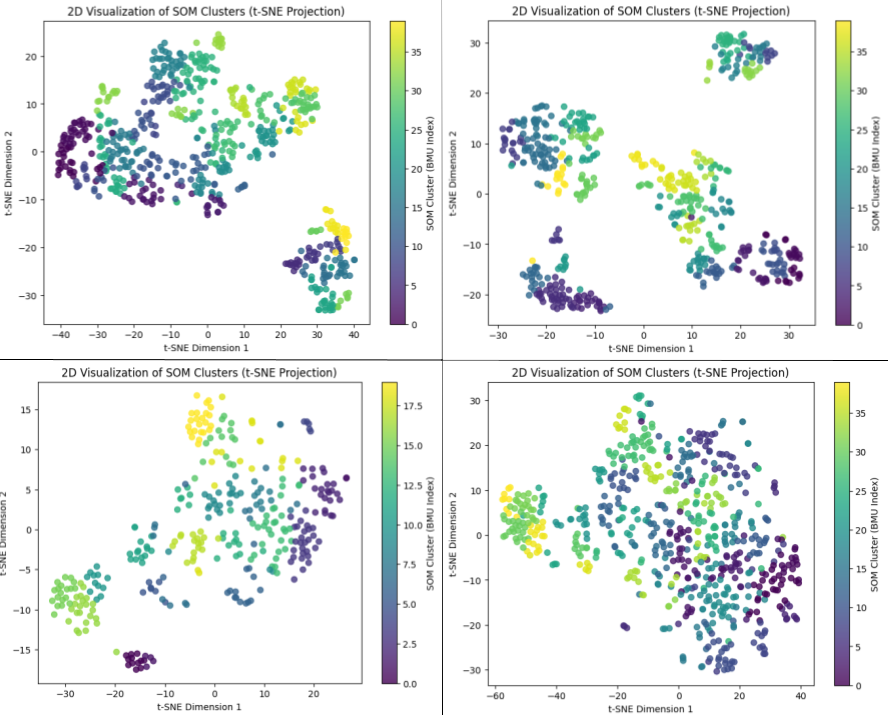} 
  \caption{Illustration of high dimensional t-SNE down sampled datapoints (question vectors) of SOM determined clusters for XYZ (top left corner), KVA (top right corner), NOG (bottom left corner) and DTK (bottom right corner).}
  \label{fig:som-intracluster}
\end{figure}
\FloatBarrier

\subsubsection{Gaussian Mixture Models (GMM)}
In this approach, the high-dimensional question embedding space is modeled as a mixture of Gaussian distributions. Instead of producing explicit clusters, the GMM provides each question with a probability distribution across multiple Gaussian components. In practice, after training a GMM on the embeddings, each question is represented by a probability vector---the posterior probabilities for belonging to each component \citep{bishop_pattern_2006}. We used these probability vectors as feature profiles for questions and determined similarity by comparing the distributions. Specifically, given a source question’s probability vector, we calculate the Kullback–Leibler (KL) divergence between that vector and every other question’s vector. A smaller KL divergence indicates that two questions have more similar probability profiles (and thus are likely to be similar). The top-N questions with the lowest KL divergences to the source are returned as recommendations (as we discuss later, this probabilistic similarity approach did not perform as well as expected in practice). We tuned GMM hyperparameters for each math subject domain (XYZ, KVA, NOG, DTK) separately, since the question formats differ significantly by subject. Key configuration details include:

\begin{description}
    \item[Number of Components] We determined the optimal number of Gaussian components for each subject using the Bayesian Information Criterion (BIC) and by examining silhouette scores on the question embeddings. This process yielded different optimal component counts per subject (16 for XYZ, 18 for KVA, 15 for NOG, and 31 for DTK). Each subject’s questions were modeled with a GMM using its respective optimal number of components (effectively capturing those many latent topic regions in the embedding space).
    \item[Covariance Type] We used full covariance matrices for each Gaussian component, allowing each distribution to capture complex shapes in the embedding space. This flexibility is crucial for modeling the intricate structure of math question embeddings, which may not be spherical or uniform.
    \item[Initialization] We employed k-means clustering initialization for the GMM’s Expectation-Maximization algorithm. This provides a robust starting point and improved convergence speed and stability, ensuring the mixture components begin in distinct regions of the embedding space.
    \item[Convergence Criteria] The GMM was trained until the change in log-likelihood fell below a tolerance of $1e-3$ or a maximum of 100 EM iterations was reached. These criteria ensured the model converged to a stable mixture solution without excessive computation.
\end{description}

This GMM method differs from the other recommendation techniques (cosine similarity and SOM) in that it does not directly compute distance between question embeddings or place questions on a fixed grid. Instead, it offers a probabilistic profile for each question that contains the probability for the question belonging to each Gaussian component (cluster). We hypothesized that this profile-based representation might capture nuanced semantic relationships not evident from a single embedding vector. However, the similarity comparison via KL divergence turned out to be challenging to get right, ultimately undermining the effectiveness of this method, which the results and following analysis will emphasize and contextualize further.

\subsection{Data Collection and Preprocessing}

User interaction data is collected from an LMS utilized for the SweSAT (or högskoleprovet, a standardized Swedish entrance exam similar to the American SAT). The data is gathered through our Django-based system, which logs detailed information at two levels:

\begin{enumerate}
    \item \textbf{Quiz Sessions}:
    Each quiz session record captures:
    \begin{itemize}
        \item \textbf{User Identification:} The unique identifier for each user (when available), enabling tracking of user-specific engagement and performance.
        \item \textbf{Algorithm Used:} The recommendation algorithm (cosine similarity, SOM, or GMM) was applied to new sessions in a randomized manner. This approach is supported by the experimental methodology described by \citet{kohavi_practical_2007}, which demonstrates that random assignment in online controlled experiments (or A/B tests) is crucial to remove biases and to attribute observed differences solely to the treatment effects. Despite the small variations in session distributions among the algorithms, the random assignment ensures an unbiased evaluation of each algorithm’s effectiveness.
        \item \textbf{Question Types}: A JSON field that stores information about the types of questions (subject categories as described in \ref{sec:recommendation-algorithms}) presented during the session.
        \item \textbf{Timestamps:} The start and end times of each session, which allow us to calculate session duration.
        \item \textbf{User Ratings:} Post-session ratings provided by users, which indicate their subjective evaluation of the recommendation quality. (Note: Not all sessions include a user rating.)
    \end{itemize}
    \item \textbf{Session Questions}:
    For each question within a quiz session, the system records:
    \begin{itemize}
        \item \textbf{Session Association:} A reference to the corresponding quiz session.
        \item \textbf{Question Identification:} A link to the specific question being answered.
        \item \textbf{Sequence Order:} The order in which the question appears within the session.
        \item \textbf{Presentation and Response Timestamps:} The exact times when the question was presented and when the answer was submitted, enabling the calculation of response time.
        \item \textbf{Correctness:} A Boolean value indicating whether the question was answered correctly.
    \end{itemize}
\end{enumerate}

Data is continuously logged as users interact with the LMS, ensuring that both macro-level session information and micro-level question responses are captured. To maintain data quality and ensure the reliability of subsequent analyses, we filter out outlier sessions for some analysis such as duration results. For example, sessions having a duration outside of the 90\% most central data are filtered out as these may indicate that a user forgot to end their session or accidentally started one and ended it immediately. For some metrics, we also excluded sessions having a duration less than 5 seconds as well as sessions without a clear duration. Furthermore, preprocessing steps include converting timestamps to standard datetime formats, computing session durations and response times, and verifying data consistency across the recorded fields. This comprehensive data forms the foundation for our analysis, allowing us to evaluate both the technical performance of our recommendation algorithms and their educational impact.

\section{Results}
In this section, we present the comprehensive evaluation of our recommendation system, which leverages deep embeddings generated by Meta’s Llama-3.2-11B-Vision-Instruct model and compares three distinct methods: cosine similarity, Self-Organizing Maps (SOM) and Gaussian Mixture Models (GMM). We begin by analyzing session-level metrics such as session count, quiz duration, and user ratings, providing quantitative insights into how users interact with the system. Next, we delve into question-level analyses by examining correctness rates and response time distributions (after filtering out extreme outliers). Finally, in section \ref{sec:streaks}, we illustrate data that shows consequences in terms of users’ wrong-answer streaks derived from the recommendations provided by each algorithm. 

\begin{figure}[htbp]
  \centering
  \includegraphics[width=0.7\linewidth]{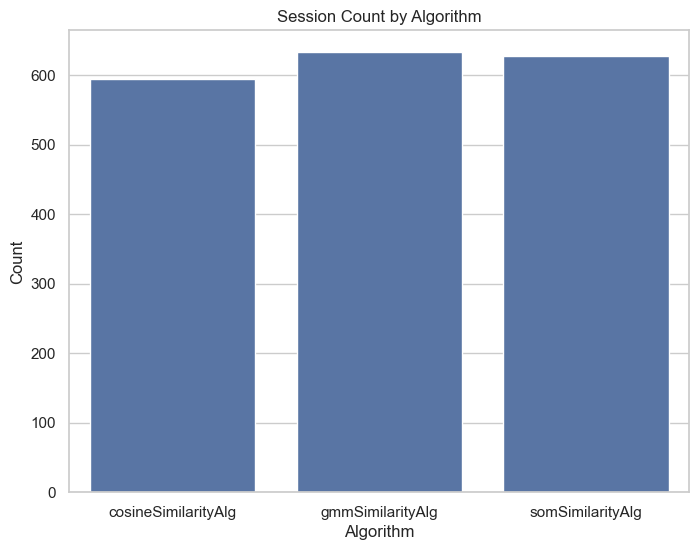} 
  \caption{A count plot shows that the study includes 1855 quiz sessions distributed among the three algorithms.}
  \label{fig:counts}
\end{figure}

\begin{figure}[htbp]
  \centering
  \includegraphics[width=0.7\linewidth]{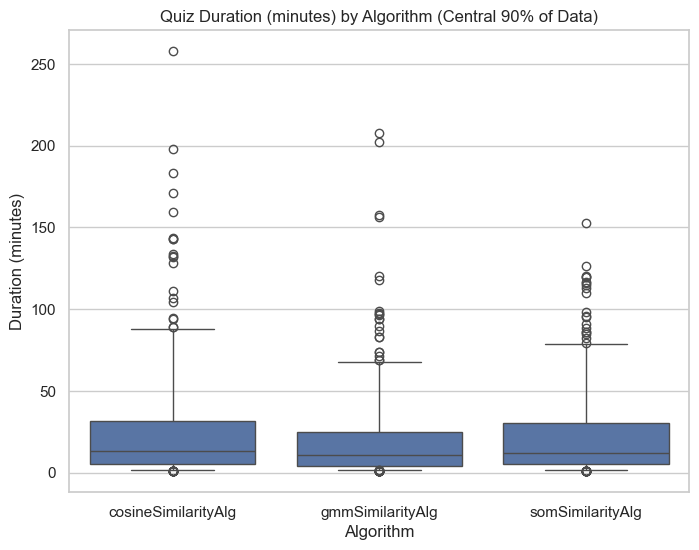} 
  \caption{Boxplots of quiz durations indicate that the central 90\% (5th to 95th percentiles) of session durations range from 0.95 to 257.73 minutes. The other 10\% of session durations are considered unreliable outliers. For instance, outlier session durations of up to 16{,}000 minutes were found, presumably because the user forgot to reset the quiz and returned days later. Note that sessions lasting less than 5 seconds were removed as outliers, and sessions without recorded durations (e.g., unfinished sessions) are not included either.}
  \label{fig:durations}
\end{figure}

\begin{figure}[htbp]
  \centering
  \includegraphics[width=0.7\linewidth]{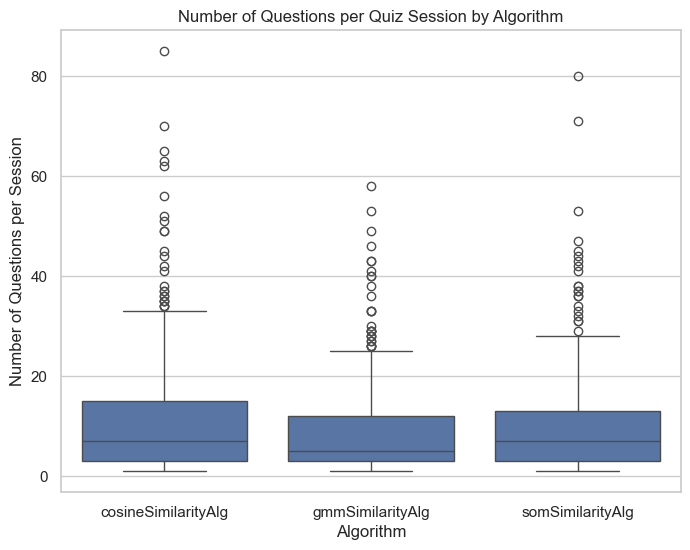} 
  \caption{This boxplot shows the distribution of the number of questions per quiz session for each recommendation algorithm. Note: Sessions lasting less than 5 seconds were removed as outliers and sessions without durations are not included either.}
  \label{fig:questions-per-algorithm}
\end{figure}

\begin{figure}[htbp]
  \centering
  \includegraphics[width=0.7\linewidth]{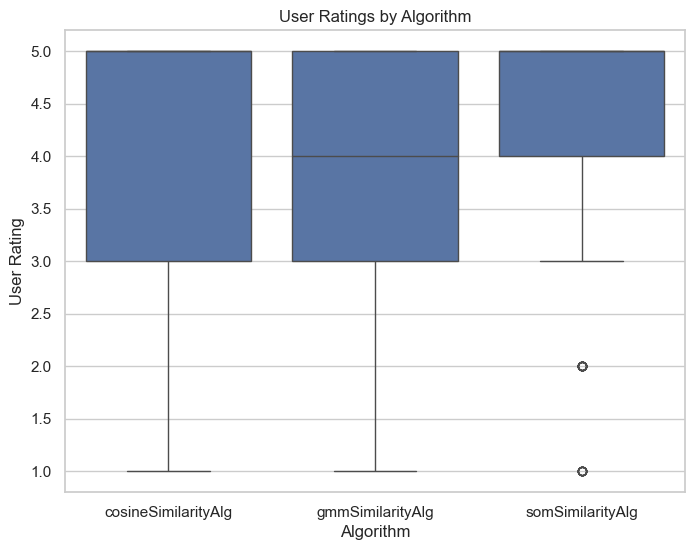} 
  \caption{Shows the distribution of user ratings on a provided scale from 1 to 5 where 5 is the best rating. Note: Users were not required to provide a rating. Consequently, there are not ratings for all sessions.}
  \label{fig:ratings}
\end{figure}

\begin{figure}[htbp]
  \centering
  \includegraphics[width=0.7\linewidth]{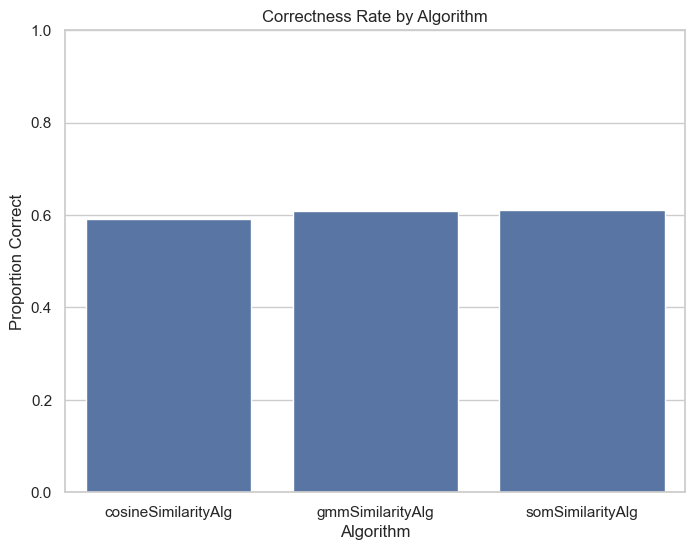} 
  \caption{A bar chart comparing correctness rates shows that the average quiz session correctness over each respective algorithm is 59.2\%, 60.9\% and 61.2\%, where a lower correctness rate directly implies more challenging recommendations.}
  \label{fig:correctness}
\end{figure}

This multi-faceted evaluation not only highlights the technical performance of the recommendation methods but also explores their educational impact. Particularly, how they balance the need for both precision and variation in challenging users. The subsequent figures (Figures \ref{fig:counts}--\ref{fig:streaks}) together with tables (Tables \ref{table:durations_table}--\ref{table:streaks}) detail these analyses and form the basis for our discussion of the effectiveness of each algorithm in enhancing math learning outcomes.

\subsection{Session-Level Analysis}

\FloatBarrier

\FloatBarrier

\begin{table}[h]
\centering
\caption{A table describing the session duration in minutes, using the same data as plotted in figure \ref{fig:durations}. Note: Outliers beyond 5th–95th percentile were removed for this summary. Sessions lasting less than 5 seconds were removed as outliers and sessions without durations are not included either, hence counts do not sum to total 1855.}
\begin{tabular}{lccccccc}
\toprule
algorithm & count & mean & std & min & 25\% & 50\% & max \\
\midrule
cosineSimilarityAlg & 367 & 25.63 & 33.86 & 0.95 & 5.33 & 13.13 & 257.73 \\
gmmSimilarityAlg & 415 & 20.18 & 26.48 & 1.02 & 4.32 & 11.08 & 207.59 \\
somSimilarityAlg & 391 & 22.33 & 26.23 & 0.96 & 5.16 & 12.32 & 152.62 \\
\bottomrule
\end{tabular}
\label{table:durations_table}
\end{table}

\FloatBarrier

\begin{table}[h]
\centering
\caption{A table representation of the data plotted in figure \ref{fig:questions-per-algorithm}. Note: Outliers beyond 5th–95th percentile \emph{were not} removed for this summary. However, sessions lasting less than 5 seconds were removed as outliers and sessions without durations are not included, hence counts do not sum to total 1855.}
\begin{tabular}{lcccccccc}
\toprule
algorithm & count & mean & std & min & 25\% & 50\% & 75\% & max \\
\midrule
cosineSimilarityAlg & 395.0 & 11.02 & 12.23 & 1.0 & 3.0 & 7.0 & 15.0 & 85.0 \\
gmmSimilarityAlg & 444.0 & 8.87 & 9.08 & 1.0 & 3.0 & 5.0 & 12.0 & 58.0 \\
somSimilarityAlg & 421.0 & 9.68 & 10.08 & 1.0 & 3.0 & 7.0 & 13.0 & 80.0 \\
\bottomrule
\end{tabular}
\label{table:questions-per-session-table}
\end{table}

\FloatBarrier

\begin{table}[h]
\centering
\caption{A table representation of the data plotted in figure \ref{fig:ratings}. Note: Users were not required to provide a rating. Consequently, there are not ratings for all sessions.}
\begin{tabular}{lcccccccc}
\toprule
algorithm & count & mean & std & min & 25\% & 50\% & 75\% & max \\
\midrule
cosineSimilarityAlg & 221.0 & 4.05 & 1.22 & 1.0 & 3.0 & 5.0 & 5.0 & 5.0 \\
gmmSimilarityAlg & 238.0 & 3.96 & 1.25 & 1.0 & 3.0 & 4.0 & 5.0 & 5.0 \\
somSimilarityAlg & 234.0 & 4.17 & 1.13 & 1.0 & 4.0 & 5.0 & 5.0 & 5.0 \\
\bottomrule
\end{tabular}
\label{table:ratings}
\end{table}

\clearpage
\subsection{Question-Level Analysis}

\FloatBarrier

\begin{figure}[htbp]
  \centering
  \includegraphics[width=1\linewidth]{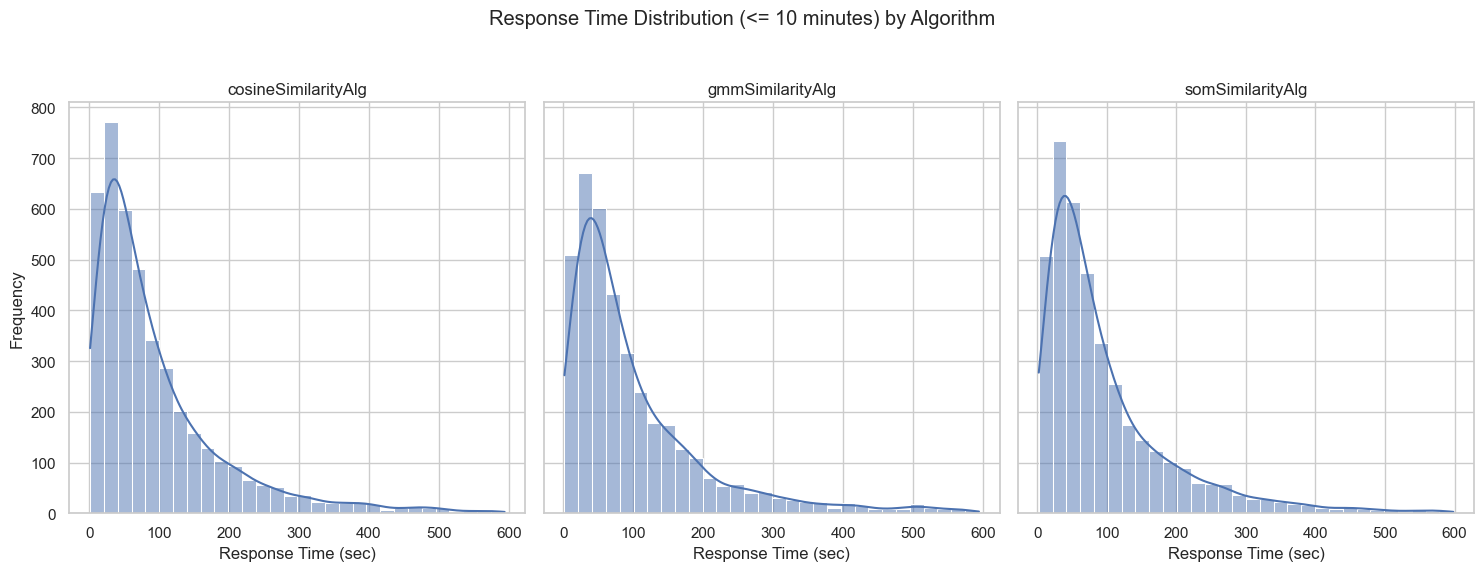} 
  \caption{The figure displays the distribution of question response times for three recommendation algorithms. For cosineSimilarityAlg, the median response time is 64.49 seconds with a mean of 95.82 seconds and a 95th percentile of 295.92 seconds. In comparison, gmmSimilarityAlg has a median of 67.19 seconds, a mean of 101.35 seconds, and a 95th percentile of 317.37 seconds, while somSimilarityAlg shows a median of 66.68 seconds, a mean of 97.86 seconds, and a 95th percentile of 296.38 seconds. Overall, the response time distributions exhibit a heavy right tail, indicating that although half of the responses are near one minute, a notable fraction extend significantly longer.}
  \label{fig:responsetimes}
\end{figure}
\FloatBarrier

\subsection{Analysis of Wrong-Answer Streaks} \label{sec:streaks}

\begin{figure}[htbp]
  \centering
  \includegraphics[width=1\linewidth]{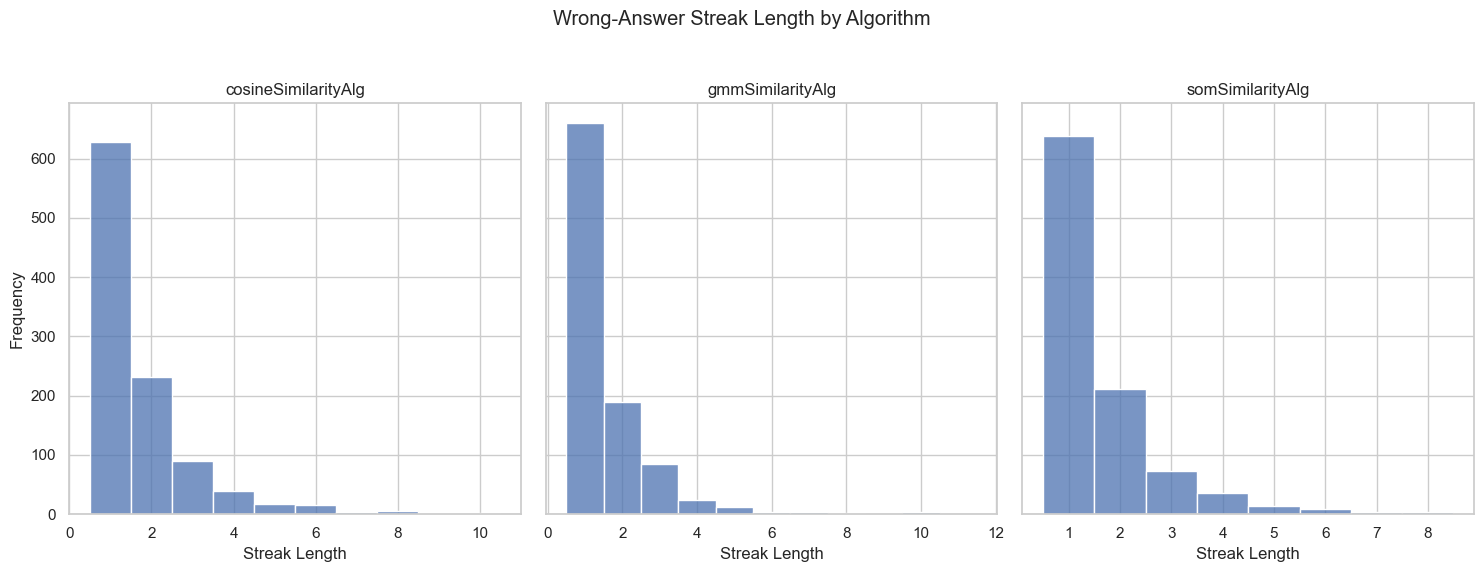} 
  \caption{A plot illustrating consecutive wrong-answer streaks within sessions.}
  \label{fig:streaks}
\end{figure}
\FloatBarrier

\begin{table}[h]
\centering
\caption{The table shows the distribution (in percentages) of consecutive wrong-answer streak lengths for each algorithm. A streak length of 1 indicates a single wrong answer before a correct one, while higher numbers (2, 3, etc.) represent longer runs of incorrect answers in a row. For all three algorithms, most streaks are length 1, but the table reveals differences in how frequently longer streaks occur. For example, \texttt{gmmSimilarityAlg} exhibits the highest proportion of single-answer streaks (67.01\%), whereas \texttt{somSimilarityAlg} has slightly fewer single-answer streaks but a higher share of streaks at lengths 4, 5 or 6. Overall, these percentages indicate that although most incorrect responses occur singly, each algorithm occasionally sees longer runs of consecutive wrong answers.}
\begin{tabular}{lcccccccccc}
\toprule
Streak length & 1 & 2 & 3 & 4 & 5 & 6 & 7 & 8 & 9 & 10 \\
\midrule
cosineSimilarityAlg & 60.68 & 22.42 & 8.60 & 3.77 & 1.64 & 1.55 & 0.39 & 0.58 & 0.19 & 0.19 \\
gmmSimilarityAlg & 67.01 & 19.29 & 8.63 & 2.44 & 1.22 & 0.41 & 0.30 & 0.20 & 0.10 & 0.30 \\
somSimilarityAlg & 64.71 & 21.40 & 7.40 & 3.65 & 1.32 & 0.81 & 0.41 & 0.30 & 0.00 & 0.00 \\
\bottomrule
\end{tabular}
\label{table:streaks}
\end{table}
\FloatBarrier

\section{Discussion}
\subsection{Overview of Key Findings}
Our study compared three distinct recommendation methods---cosine similarity, Self-Organizing Maps (SOM), and Gaussian Mixture Models (GMM)---using engagement metrics such as session duration, question count per session, and user ratings, as well as performance indicators like correctness rates and wrong-answer streaks. The results indicate that while both cosine similarity and SOM yield robust engagement, GMM underperforms on almost all metrics. Cosine similarity delivers highly similar question matches, SOM offers a beneficial mix of similarity and novelty, and GMM, implemented as a probability-based similarity approach using KL divergence, falls short in maintaining user interest (Tables \ref{table:durations_table}--\ref{table:ratings}).

It is noteworthy that although the mean session duration for the cosine similarity algorithm is slightly higher (as shown in Table \ref{table:durations_table}), this metric is sensitive to occasional long sessions that may not reflect typical user behavior: its longest session is approximately 258 minutes, compared to around 153 minutes for SOM. Despite not having the highest mean session duration, the SOM-based approach shows a more consistent engagement profile across multiple metrics, including higher user ratings and a balanced distribution of wrong-answer streaks. Thus, while cosine similarity occasionally produces longer sessions in terms of time or number of questions, SOM appears to foster a more uniformly engaging learning experience.

This pattern aligns with prior research in adaptive learning systems; for example, \citet{lim_measuring_2022} validated a comprehensive learner satisfaction questionnaire and demonstrated that enhanced learner satisfaction is closely tied to sustained system use and overall educational success. In this context, our findings reinforce the view that both the precision of content matching and the inclusion of varied, engaging recommendations are critical for achieving long-term user engagement.

\subsection{Cosine Similarity: Precision at a cost}
The cosine similarity method excels at retrieving questions that are nearly identical to the query item, ensuring high conceptual relevance. This precision can be advantageous in reinforcing specific problem types, particularly after a student makes an error. However, the method’s strict focus on similarity may lead to overly repetitive recommendations. Our results show that sessions using cosine similarity tend to have the lowest correctness rates, wrong-answer streaks distributed higher over streak lengths longer than 1 compared to how the streaks were distributed across lengths for SOM and GMM (Figure \ref{fig:streaks} and Table \ref{table:streaks}), lower ratings and lower overall engagement compared to SOM (except for the reliable data in Figure \ref{fig:questions-per-algorithm} and Table \ref{table:questions-per-session-table}, which indicate that cosine similarity achieved longer average quizzes in terms of questions per session). Interestingly, the cosine method also registers the lowest median and mean question response times (Figure \ref{fig:responsetimes}), suggesting that users might be more confident in solving questions that are nearly identical to those immediately preceding them. These outcomes suggest that while cosine similarity is effective in delivering highly relevant content, its lack of diversity might lead to learner fatigue and frustration over time. Similar to the observations reported by \citet{ibrahim_revisiting_2025}, our results underscore that an excessive reliance on strict similarity can limit recommendation diversity and contribute to user disengagement \citep{anwar_filter_2024, zhao_fairness_2023}.

\subsection{SOM: Balancing Similarity and Novelty}
In contrast, the SOM‐based approach projects high-dimensional question embeddings onto a two-dimensional grid, which inherently clusters conceptually similar queries while also “spacing” them apart to introduce controlled novelty. This duality not only ensures that recommendations remain aligned with the original query but also injects enough diversity to engage students more effectively. For instance, \citet{bhaskaran_design_2021} developed a cluster‐based intelligent hybrid recommender system for e-learning that partitions learners and learning resources into clusters based on their individual profiles and learning styles. Their experiments on a dataset of 1000 learners showed that this clustering approach led to improved predictive accuracy (as demonstrated by enhancements in Recall and Ranking Score metrics) as well as higher user satisfaction and an increase in the number of completed lessons, all while reducing computational time. These outcomes suggest that leveraging clustering to tailor recommendations can effectively optimize system performance in personalized educational settings.

\subsection{GMM: Probability-Based Similarity and Underperformance}
We initially adopted GMM to model the embedding space as a mixture of Gaussian distributions, believing it would capture nuanced semantic relationships. However, in our actual implementation, we used the GMM output to generate probability vectors for each question, each element representing the probability of belonging to a given Gaussian component and compared these vectors via KL divergence to find “similar” questions. Empirically, this approach did not align well with user-perceived similarity. GMM-based recommendations led to shorter sessions, fewer questions per session, and lower user ratings. These findings suggest that while a probabilistic profile can theoretically capture rich information about a question’s latent concepts, KL divergence was not an effective measure of similarity in this context. It may have been overly sensitive to small probability differences or poorly correlated with true pedagogical similarity. Indeed, \citet{nielsen_guaranteed_2016}, showed that even guaranteed bounds on the KL divergence for univariate mixtures can be highly sensitive to slight variations in component probabilities, which may lead to significant discrepancies in similarity measurements.

\subsection{Comparative Insights and Educational Implications}
The divergent outcomes of these methods underscore the trade-offs inherent in educational recommender systems. Cosine similarity offers precise content matching but risks repetitive question sets. SOM effectively balances similarity and novelty, improving user engagement through more varied recommendations. GMM, while promising in principle, underperformed due to the way similarity was computed. These insights highlight the importance of matching both the representation (e.g., probabilistic profiles) and the similarity metric (e.g., KL divergence vs. direct distance) to the learning context. A more cluster-oriented usage of GMM, filtering questions by their most probable component and then applying a direct distance measure, may better harness GMM’s latent-structure capabilities.

\subsection{Limitations and Future Directions}
\begin{figure}[htbp]
  \centering
  \includegraphics[width=1\linewidth]{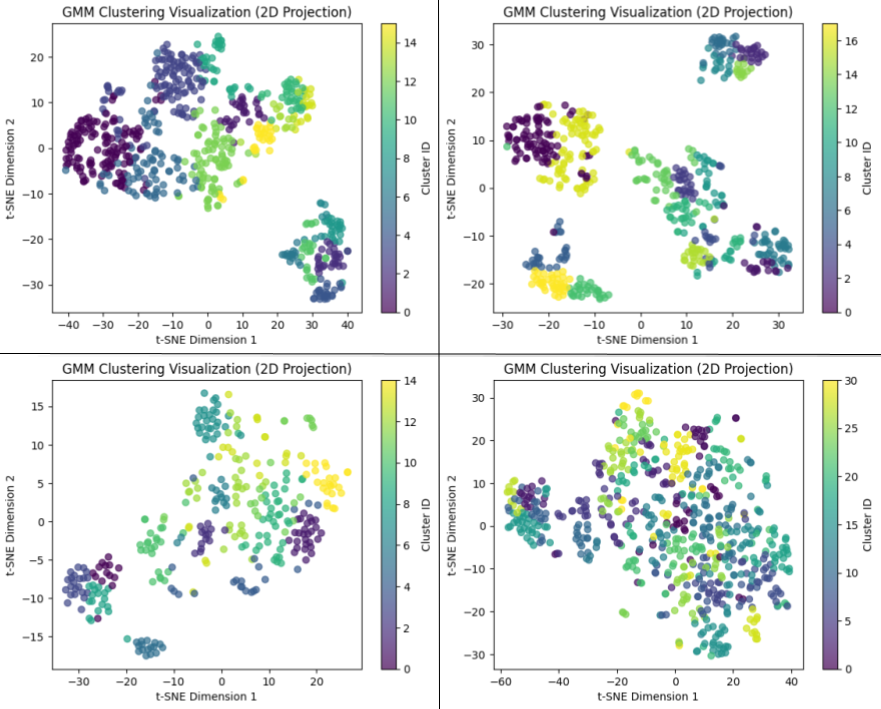} 
  \caption{A plot illustrating how the GMM as a clustering method determined components (clusters). It might be useful for future work because it demonstrates how GMM clustered our data, which is comparable to the clustering achieved with SOM.}
  \label{fig:gmm-plot}
\end{figure}

Although session duration, question count, and user ratings offer insights into learner engagement, they do not directly measure long-term knowledge gains. Future work should incorporate direct assessments of learning outcomes, such as pre- and post-tests, to evaluate how each recommendation method contributes to actual skill improvement. In parallel, refining the GMM approach by either:

\begin{enumerate}
    \item Replacing KL Divergence with a more suitable distributional metric (e.g., Jensen--Shannon divergence), or
    \item Reverting to a True Clustering Paradigm by assigning each question to its top component and using a direct distance measure within that subset
\end{enumerate}

could potentially improve performance if the latent features are preserved more faithfully. For the second proposal of how future work might refine our GMM approach by reverting to a GMM clustering approach, we have experimented and plotted preliminary clusters that our GMM models created (Figure \ref{fig:gmm-plot}). At first glance, these clusters seem to have potential for better performance than the SOM-based clusters through clear and well-separated clusters, which might not be surprising considering our efforts to find the optimal number of components (clusters) with BIC and silhouette scores as described in Section \ref{sec:recommendation-algorithms}.

\FloatBarrier

Exploring hybrid models that combine the precision of cosine similarity with the diversity of SOM, or a better-tuned GMM, remains an attractive direction. By balancing reinforcement of familiar concepts with the introduction of novel challenges, educational recommender systems can maintain student interest, reduce frustration, and potentially improve long-term learning outcomes.

\section{Conclusion}
Our evaluation of three recommendation methods, cosine similarity, Self-Organizing Maps (SOM), and a probability-based Gaussian Mixture Model (GMM), reveals distinct trade-offs in enhancing math learning within an LMS. The cosine similarity method reliably retrieves highly relevant questions but may hinder sustained engagement due to its repetitive nature. The SOM-based approach strikes a more effective balance between precision and novelty, resulting in longer session durations, higher user ratings, and improved overall engagement. GMM, while conceptually promising for modeling latent question characteristics, consistently underperformed in our study because KL divergence did not effectively capture meaningful similarity in practice. These findings underscore the critical importance of incorporating both robust representations and appropriate similarity metrics into educational recommender systems. Balancing the reinforcement of familiar concepts with the introduction of novel challenges appears essential for maintaining student interest and optimizing learning outcomes. Future work should explore hybrid or alternative methods that better leverage GMM’s latent structure, potentially by using a cluster-oriented approach or improved similarity measures, alongside direct assessments of learning gains. Ultimately, our study reinforces that personalized learning systems benefit from a nuanced recommendation design: one that not only emphasizes content relevance but also fosters a dynamic and engaging learning environment.

\bibliographystyle{apalike}  
\bibliography{references}

\end{document}